\documentclass[11pt]{article}

\usepackage[final]{acl}
\usepackage{amsmath}   
\usepackage{amsfonts} 
\usepackage{latexsym}
\usepackage{booktabs}
\usepackage{graphicx}
\usepackage{subcaption}
\usepackage[T1]{fontenc}

\usepackage[utf8]{inputenc}

\usepackage{microtype}
\usepackage{pgfplots}
\usepackage{pgfplotstable}
\pgfplotsset{compat=1.18}
\usepackage{inconsolata}

\usepackage{color}
\usepackage{algorithmic}
\usepackage[inline]{enumitem}
\usepackage[switch]{lineno}

\usepackage{mathtools}
\usepackage{multirow}
\usepackage{url}
\usepackage{tabto}
\usepackage[most]{tcolorbox}

\newtcolorbox{promptbox}[1]{
    colback=gray!5!white,
    colframe=gray!75!black,
    fonttitle=\bfseries,
    title=#1,
    arc=2mm,
    boxrule=0.5pt,
    left=10pt,
    right=10pt,
    top=5pt,
    bottom=5pt
}
\usepackage{fontspec}

\usepackage{polyglossia}
\setmainlanguage{english}
\setotherlanguages{hindi,bengali,thai}
\newfontfamily\devanagarifont[Script=Devanagari, Path=./]{NotoSerifDevanagari.ttf}

\newfontfamily\bengalifont[Path=./, BoldFont={Kalpurush.ttf}, BoldItalicFont={Kalpurush.ttf}]{Kalpurush.ttf}

\newfontfamily\thaifont[Script=Thai, Path=./]{NotoSerifThai-Regular.ttf}

  \XeTeXlinebreaklocale "th"

\title{When Meaning Travels: A Granular Lens on Hybrid-MoE's Role in Idiomatic Understanding for Language Models}
\author{
        \textbf{Sarmistha Das}$^{1*}$,\quad
        \textbf{Vaibhav Vishal}$^{1*}$,\quad
        \textbf{Shreyas Guha}$^{1*}$,\quad
        \textbf{Amaan Ali}$^{1*}$,\\
        \textbf{Kitsuchart Pasupa}$^2$, \quad\textbf{Sriparna Saha}$^1$\\
        $^1$Department of Computer Science and Engineering, Indian Institute of Technology Patna, India\\
        $^2$School of Information Technology, King Mongkut's Institute of Technology Ladkrabang, Thailand\\
\texttt{\{sarmistha1515, vvaibhav728, shreyas.slg, ali2003.amaan\}@gmail.com}\\
\texttt{kitsuchart@it.kmitl.ac.th, sriparna@iitp.ac.in}
  }

\begin{document}
\maketitle
\begin{abstract}
In the contemporary epoch of multilingual education, learning idioms provides a fascinating gateway towards creativity, cultural values, historical context, and diverse perspectives inherent to various linguistic traditions. This paper showcases the navigation of retaining figurative and cultural semantics in low-resource Southeast Asian languages such as Hindi, Bengali, and Thai, where culturally rich idioms pose significant obstacles for computational modeling and cross-linguistic transfer due to their deep metaphorical complexity. To tackle such complexity, we present \textit{Varnika} 
\begin{hindi}(वर्णिका)\end{hindi}, 
a reconstructed multimodal idiom corpus comprising 3,533 multilingual idioms, enriched with seven idiomatic tones aligned with both textual and visual representations. Additionally, to infer informative idiomatic understanding, we introduce a Hybrid Mixture-of-Experts (HybridMoE) framework that embeds multiple idiomatic expert opinions while mitigating expert sparsity by integrating outputs from both selected and unselected experts through controlled hybridization, further augmented with Idiomatic Property Signals via masked multimodal embeddings. To analyze the performance across multiple dimensions, we propose the \textit{IDIO-TONE} and Idiomatic Validation Score, a three-stage evaluation pipeline measuring (i) literal translation fidelity, (ii) visual-semantic alignment, and (iii) idiomatic meaning retention. Empirical evaluations highlight that HybridMoE achieves 5--6\% performance gains across advanced vision language models, demonstrating improved representation of figurative language and culturally embedded meaning in multilingual multimodal settings\footnote{Resources are available at \url{https://github.com/sarmistha-D/Hybrid_MOE}.\\
$^*$ These authors contributed equally.}.
\end{abstract}

\section{Introduction}
Idioms constitute a pivotal conduit for expressive language, encapsulating the figurative essence of human experience, culture, history, and creativity. Cross-linguistic idiom learning plays a critical role in enhancing language proficiency, deepening cultural literacy, and strengthening both semantic awareness and affective engagement with language~\cite{liontas2017teach}.  Despite rapid advances in large language models and their growing integration into educational and social ecosystems~\cite{kasneci2023chatgpt,nayeem2024kidlm,rong2024kids,maji2025sanskriti,maji2025drishtikon}, accurately decoding culture-specific lexical patterns and capturing figurative metaphors grounded in linguistic convention remains a persistent challenge~\cite{liu2025cultural}. Languages across South and Southeast Asia, including Hindi, Bengali, and Thai, exhibit shared phonological traits and deeply rooted cultural motifs, often giving rise to semantically parallel idiomatic expressions. For example, the Hindi idiom \begin{hindi}एक पत्थर से दो शिकार\end{hindi} (\textit{ek patthar se do shikār}, ``two outcomes from a single action'') finds close counterparts in Bengali \begin{bengali}এক পাথর দুই পাখি মারা\end{bengali} (\textit{ek pāthor dui pākhi mārā}) and Thai \begin{thai}ยิงปืนนัดเดียวได้นกสองตัว\end{thai} (\textit{ying peun nát diao dâi nók sǒng tua}). Nevertheless, idioms fundamentally resist compositional interpretation; their meanings are intrinsically tied to cultural grounding, shared world knowledge, and conventionalized contextual usage~\cite{sporleder2009unsupervised,fornaciari2024hard}.
\begin{figure*}[t]    
    \centering
\includegraphics[width=0.8\textwidth]{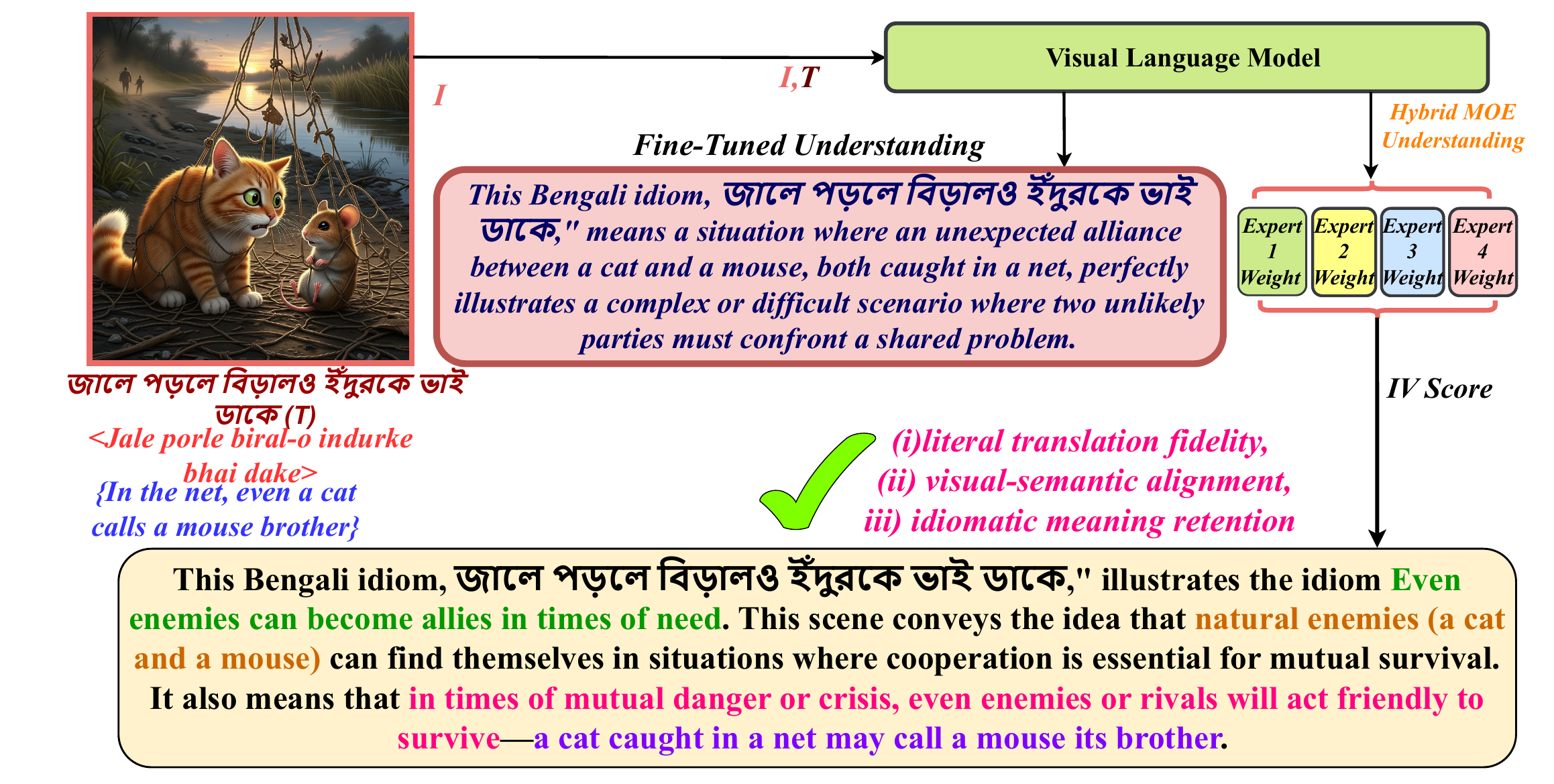}
    \caption{Visual Representation of Idiomatic Understanding via Hybrid Mixture-of-Experts.}
    \label{fig:intro}
\end{figure*}
Consequently, figurative language understanding has emerged as a central research theme in contemporary Natural Language Processing (NLP), with substantial progress in related tasks such as metaphor identification~\cite{dagan2005pascal,gao2018neural,chakrabarty2021mermaid}, simile detection~\cite{niculae2014brighter,mpouli2017annotating,zeng2020neural}, pun recognition~\cite{poliak2018collecting}, and idiom retrieval~\cite{tan2016neural,lee2016quote}. Despite these advances, a unified and comprehensive evaluation framework for idiomatic interpretation remains largely absent. FLUTE~\cite{chakrabarty2022flute} introduced a large-scale figurative language inference benchmark spanning sarcasm, simile, metaphor, and idiom understanding, while its multimodal extension, V-FLUTE~\cite{saakyan2024v}, further demonstrated the critical role of visual--linguistic grounding in robust idiom comprehension.

However, both fall short in addressing the multilingual and cultural complexities of low-resource languages such as  Thai, Bengali, and Hindi, resulting in a persistent gap in figurative understanding across languages~\cite{haagsma2020magpie}. While scaling enhances model performance, dense architectures remain computationally expensive. Mixture-of-Experts (MoE) frameworks~\cite{shazeer2017outrageously} offer a solution by activating only a small subset of expert modules for each input~\cite{artetxe2021efficient,du2022glam}, making it possible to scale efficiently. MoE routes tokens through a small subset of experts, reducing computational load while preserving performance. In pragmatically rich tasks such as idiom explanation, relying solely on top-k expert selection may constrain expressiveness. Leveraging HyperNetworks~\cite{zhao2023prototype} enables cross-expert knowledge sharing, enhancing task-specific representation and semantic transfer as depicted in Figure~\ref{fig:intro}. However, preserving multimodal information in idioms, where the literal and intended meanings are semantically divergent, remains a challenging task.

Despite the growing interest in idiomatic understanding, cross-lingual research that jointly incorporates visual representations and idiomatic tones remains limited, particularly for culturally rich languages such as Thai, Bengali, and Hindi (Table~\ref{tab:relatedwork}). This gap is critical, as idioms encapsulate nuanced intercultural meaning and social cognition, yet no existing dataset provides visually grounded idioms suitable for both educational and cross-modal learning. To address this, we introduce \textit{Varnika} \begin{hindi}(वर्णिका)\end{hindi}, a multimodal idiom corpus with seven idiomatic tones. Building on this dataset, we propose a Hybrid Mixture-of-Experts (HybridMoE) framework that integrates expert-specific modules with a globally shared hypernetwork to capture diverse idiomatic properties while enabling positive expert transfer. To further enhance contextual fidelity, HybridMoE is augmented with an Idiomatic Property Signal (IPS) that selectively conditions cross-modal embeddings, and an Idiomatic Validation (IV) score that systematically evaluates literal fidelity, visual–semantic alignment, and idiomatic meaning retention. Additionally, we propose \textit{IDIO-TONE}, a label-based quality assurance and performance metric for emotional and stylistic consistency of idioms.

\begin{table*}[t]
\centering
\caption{Comparison of the proposed \textit{Varnika} dataset with leading existing resources. Abbreviations: E (English); ML (Multilingual: Thai, Hindi, and Bengali). Multimodality indicates the inclusion of paired text–image data.}
\label{tab:relatedwork}
 \resizebox{\textwidth}{!}{
\begin{tabular}{lcccccc}
\toprule
Corpus Name                                         & \multicolumn{1}{c}{Count} & Language & Explanations & Multimodality & Tone & Idiomatic Part \\ \midrule

SemEval-2013~\cite{korkontzelos2013semeval} & 4,350                                   & E           & $\times$    & $\times$  & $\times$  & $\times$              \\
FLUTE~\cite{chakrabarty2022flute} & 8,962                                   & E           & $\checkmark$    & $\times$  & $\times$  & $\checkmark$(497)               \\
V-FLUTE~\cite{saakyan2024v} & 6,027                                   & E & $\checkmark$ & $\checkmark$  & $\times$ &  $\checkmark$(370)                \\
MAGPIE~\cite{haagsma2020magpie} & 56,622                                   & E & $\checkmark$ & $\checkmark$  & $\times$ & $\checkmark$(1,756)                 \\
IRFL~\cite{yosef2023irfl} & 6,027                                   & E & $\times$ & $\checkmark$  & $\times$ & $\checkmark$(628)                 \\
\textit{Varnika} \begin{hindi}(वर्णिका)\end{hindi}(Proposed)                        & 3,533                                    & ML            & $\checkmark$                     & $\checkmark$  & $\checkmark$   & 3,533              \\ \bottomrule
\end{tabular}
}
\end{table*}

The research objectives of the current work are as follows:
\begin{enumerate}[label=(\roman*)]
\item Evaluate how Vision-Language Models (VLMs) internalize culturally grounded idiomatic knowledge, and analyze the impact of HybridMoE with IPS integration on enhancing fine-grained cultural inference and interpretative accuracy.
\item Assess the cross-lingual and cross-architectural generalization capabilities of both the proposed dataset and models, evaluating their ability to adapt to diverse linguistic settings and unseen idiomatic constructs.
\end{enumerate}

The primary contributions of this work are as follows:
\begin{enumerate}[label=(\roman*)]
\item The proposal of HybridMoE, a multimodal hypernetwork-based framework equipped with IPS designed to enhance reasoning over idiomatic constructs, alongside the formulation of the novel Idiomatic Validation (IV) and \textit{IDIO-TONE} metric to quantify idiomatic understanding.
\item The design and execution of two core tasks: (a) multimodal idiomatic interpretation leveraging state-of-the-art VLMs, and (b) evaluation of the HybridMoE mechanism for performance enhancement, aimed at benchmarking the effectiveness of the proposed dataset and model design.
\item We reconstruct the existing multimodal idiom dataset \textit{Mediom}~\cite{das2026meaningisntliteralexploring} by augmenting each instance with fine-grained idiomatic tonal annotations, and present the resulting resource as \textit{Varnika} \begin{hindi}(वर्णिका)\end{hindi}. This enhanced dataset targets low-resource languages such as Hindi, Bengali, and Thai and incorporates seven pragmatic idiomatic tones: \textit{Humor}, \textit{Ridicule}, \textit{Affection}, \textit{Aspiration}, \textit{Fear}, \textit{Sorrow}, and \textit{Deception}. 
\end{enumerate}

\section{Formulation of \textit{Varnika} \begin{hindi}(वर्णिका)\end{hindi} dataset.}

\subsection{Dataset Collection}
We reconstruct \textit{Mediom}~\cite{das2026meaningisntliteralexploring}, a multilingual, multimodal idiom dataset comprising 3,533 expressions across Hindi, Bengali, and Thai. The dataset is curated from culturally grounded and linguistically diverse sources, including online repositories, literary compilations, and archival linguistic resources. Specifically, Hindi idioms are sourced from \textit{The Simple Help}\footnote{\url{https://thesimplehelp.com/hindi-idioms-with-meanings-and-sentences}}, Bengali idioms from \textit{Bangla Probad}\footnote{\url{https://archive.org/details/in.ernet.dli.2015.455639/page/n557/mode/2up}}, and Thai idioms from a comprehensive collection~\cite{udomporn20145000}, ensuring broad cultural coverage and authenticity across the three languages.

While \textit{Mediom} provides broad multilingual and multimodal coverage, it does not explicitly capture idiomatic tonality, a key pragmatic dimension that often governs how idioms are interpreted, expressed, and visually grounded. This aspect is particularly important because idiomatic meaning extends beyond literal semantics to encode affective, social, and culturally situated intent. To address this gap, we reconstruct the dataset by augmenting each entry with a taxonomy of seven pragmatic tone labels: Humor, Ridicule, Affection, Aspiration, Fear, Sorrow, and Deception. This refinement yields a richer and more nuanced resource for studying idiomatic understanding at the intersection of language, culture, and visual representation.

\textit{Mediom} corpus spans a wide range of idiomatic constructions, including fixed expressions (e.g., \begin{thai}น้ำท่วมปาก\end{thai}, \textit{nam thuam pak}, unable to speak out), which often convey fear, hesitation, or emotional suppression; semi-fixed idioms (e.g., \begin{hindi}आसमान से गिरे, खजूर में अटके\end{hindi}, \textit{āsmān se gire, khajūr meṁ aṭke}, out of the frying pan into the fire), which encode sorrow, frustration, or helplessness; verb-object constructions (e.g., \begin{bengali}ঘি না পেয়ে নাক কাটা\end{bengali}, \textit{ghi nā peyē nāk kāṭā}, to overreact over a loss), which frequently evoke ridicule or social mockery; adjective-noun phrases (e.g., \begin{thai}ฝันหวาน\end{thai}, \textit{fan wān}, sweet dreams), which reflect affection, tenderness, and aspiration; prepositional idioms (e.g., \begin{bengali}কানে কানে বলা\end{bengali}, \textit{kāne kāne balā}, to whisper), which may express deception, secrecy, or intimate affection depending on context; and binomial formations (e.g., \begin{hindi}धीरे धीरे\end{hindi}, \textit{dhīre dhīre}, step by step), which convey gradual progress, patience, and aspiration. More broadly, several curated idioms instantiate humor through playful exaggeration, ridicule through sarcastic social commentary, fear through cautionary imagery, sorrow through expressions of loss and resignation, and deception through motifs of concealment, misdirection, and duplicity.

Motivated by the need to capture these pragmatically salient dimensions, we reconstruct \textit{Mediom}~\cite{das2026meaningisntliteralexploring} into a more expressive and semantically enriched resource, which we name \textit{Varnika} \begin{hindi}(वर्णिका)\end{hindi}. Derived from the notion of varnan (description or depiction), \textit{Varnika} is designed to foreground not only the figurative semantics of idioms but also their tonal, cultural, and affective portrayals, thereby yielding a more comprehensive multimodal representation of idiomatic meaning.

\begin{figure*}[t]    
    \centering
\includegraphics[width=0.9\textwidth]{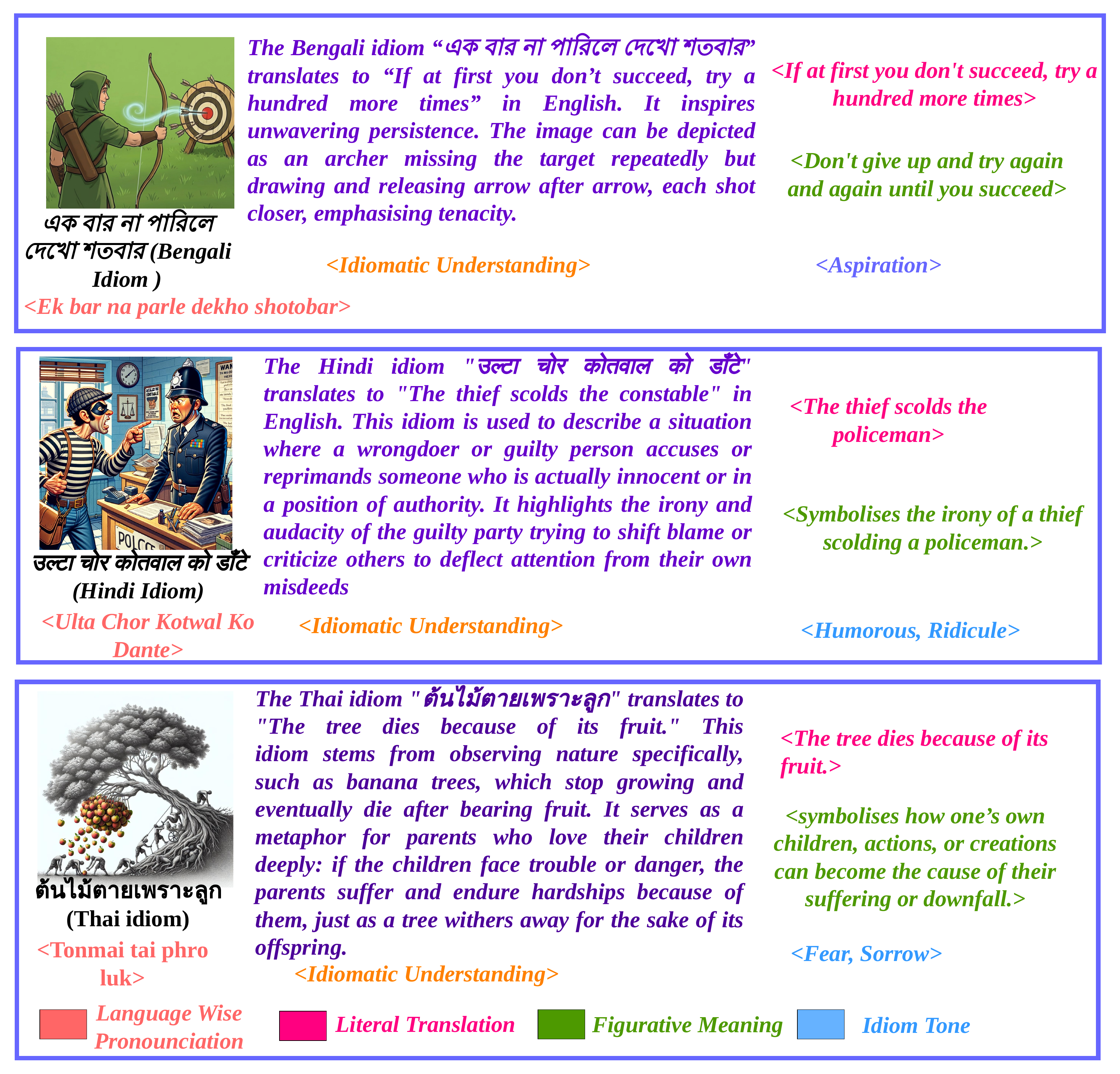}
    \caption{Sample instances of our proposed \textit{Varnika} \begin{hindi}(वर्णिका)\end{hindi} dataset.}
    \label{fig:datasample}
\end{figure*}

Syntactically flexible idioms were normalized by standardizing verb inflections and pronouns, whereas rigid forms were retained in their original surface realization to preserve cultural authenticity. A representative view of \textit{Varnika} is presented in Figure~\ref{fig:datasample}.

\subsection{Data Quality Assurance}
To ensure annotation quality and reliability, each instance underwent a two-stage validation process. First, annotations were peer-reviewed by at least two additional fluent annotators to minimize subjective bias and enhance consistency. To further assess cross-modal pragmatic alignment, we quantified the overlap between tone labels assigned from textual interpretations and those derived from corresponding visual representations, yielding an overlap of 54.75\%, indicating strong alignment between idiomatic tonal intent and visual grounding (Please refer to Appendix~\ref{dataset_quality}). Subsequently, a panel of cultural and linguistic experts performed a comprehensive final validation, evaluating each instance against the predefined tonal alignment criteria and assigning scores based on the degree of compliance. The overall annotation process achieved a substantial inter-annotator agreement of 0.65 (Cohen's kappa~\cite{gwet2014handbook}), demonstrating a high level of consistency and reliability in the resulting dataset.

\section{Methodology}
Given an idiom instance represented as a multimodal pair \((x_t, x_i)\), where \(x_t\) denotes the textual idiom expression and \(x_i\) corresponds to its associated visual depiction, this study proposes a unified framework for idiomatic interpretation through a HybridMoE architecture coupled with an IV (Idiomatic Validation) mechanism. The dataset comprises two aligned sets: textual explanation pairs \(\mathcal{D}_t = \{(x_{t_j}, y_j)\}_{j=1}^{N}\) and image explanation pairs \(\mathcal{D}_i = \{(x_{i_j}, y_j)\}_{j=1}^{N}\), where \(N\) denotes the total number of idioms and \(y_j\) represents the idiomatic meaning. 

\begin{figure*}[t]    
\centering
   \includegraphics[width=\textwidth]{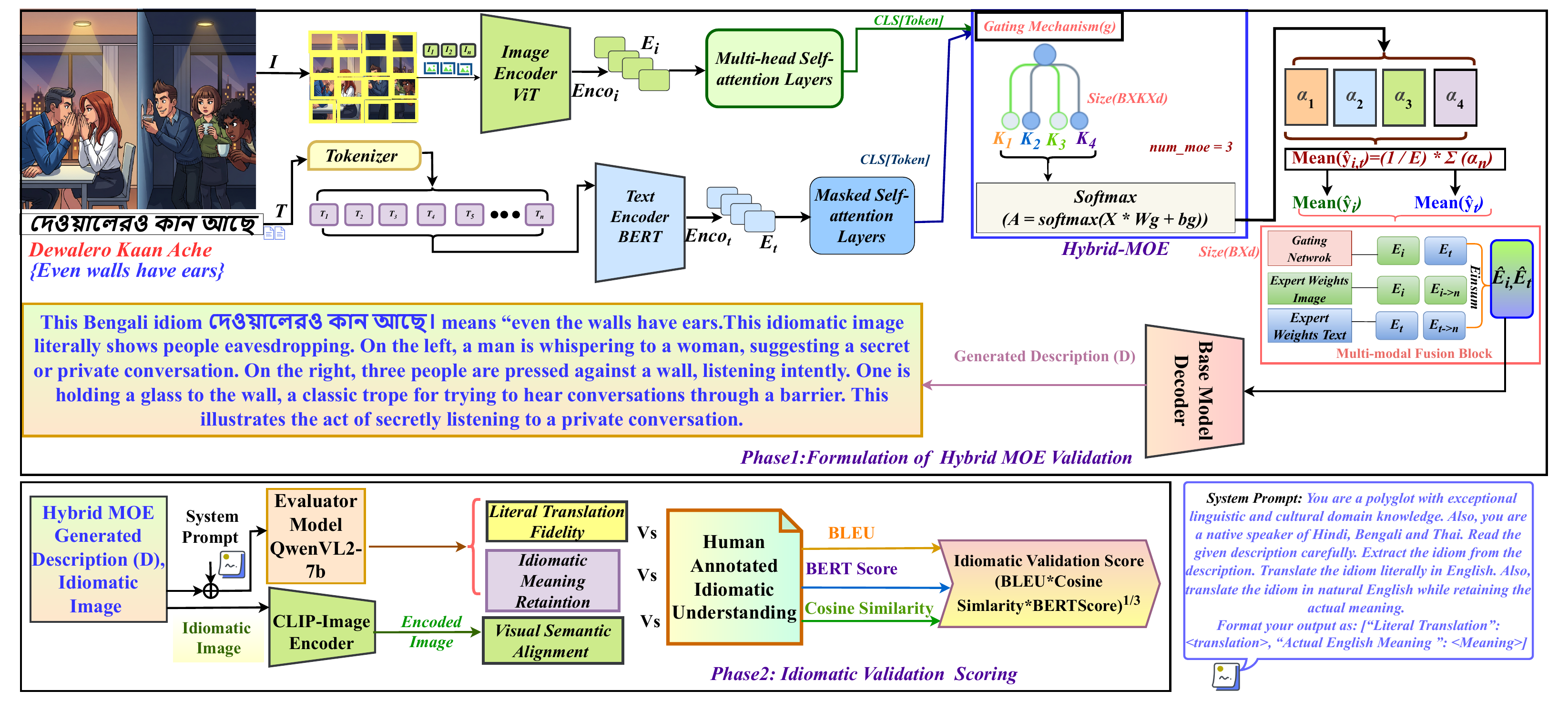}
    \caption{Architectural Viewpoint of Proposed Multimodal HybridMoE Model.}
    \label{archi_Hybrid_MOE}
\end{figure*}

The proposed framework operates in two sequential stages, as illustrated in Figure~\ref{archi_Hybrid_MOE}. 
The first stage focuses on idiom interpretation, employing HybridMoE to model and integrate semantic features from both textual and visual cues. The second stage introduces the IV (Idiomatic Validation), a tri-level evaluation mechanism designed to assess idiomatic understanding. 

\subsection{Formulation of HybridMoE}
Given an idiomatic expression \( T \) and its corresponding image \( I \), the proposed framework begins by encoding \( I \) using a ViT-patch16 based vision encoder~\cite{dosovitskiy2020image} to obtain visual embeddings \( E_i \), and encoding \( T \) with a BERT-based text encoder~\cite{devlin2019bert} to produce textual embeddings \( E_t \). The resulting hidden representations are denoted as \( H_i \in \mathbb{R}^{B \times N \times d} \) and \( H_t \in \mathbb{R}^{B \times L \times d} \), where \( B \) is the batch size, \( N \) the number of image patches, \( L \) the tokenized text length, and \( d \) the hidden dimensionality. From these, the CLS tokens are extracted via attention mechanisms as $E_i$ and $E_t$ each in \( \mathbb{R}^{B \times d} \). To encode idiomatic specificity, the CLS tokens are enriched via IPS, capturing idiom-aware visual and textual cues (see Figure~\ref{archi_Hybrid_MOE}), replacing the original embeddings. The IPS-enhanced embeddings are independently propagated through \( M = 3 \) parallel Hybrid-MoE modules. Each module consists of \( K = 4 \) expert networks with a shared architecture \( \mathbb{R}^d \rightarrow \mathbb{R}^{1024} \rightarrow \mathbb{R}^d \), and a trainable gating function defined as \( g(E) = \mathrm{Softmax}(E W_g + b_g) \in \mathbb{R}^{B \times 4} \), which dynamically computes expert weights \( \alpha_1, \ldots, \alpha_4 \). Within each block, the expert outputs are aggregated as \( \mathbf{E} = [K_i(E)]_{i=1}^{4} \in \mathbb{R}^{B \times 4 \times d} \), and further refined via a mean-pull strategy to yield an enhanced representation as \( \mathrm{Mean}(\hat{y}_{i,t}) \) and subsequently passed to a fusion layer for final integration.\\
In the fusion block, we employ Einstein summation (EinSum) to compute a weighted aggregation of the expert outputs. Specifically, each expert's adjusted output is multiplied by its corresponding gating weight and summed across the \( K \) experts to produce a single fused vector \( \hat{y} \in \mathbb{R}^{B \times d} \) per Hybrid-MoE module. Upon obtaining the outputs from all \( M \) parallel modules, we perform an element-wise averaging to derive a unified representation. This aggregated representation is then combined  \{$\hat{E}_t, \hat{E}_i$\}, effectively integrating modality-specific idiomatic signals. The fused embedding \( \hat{E}_{\text{fused}} \) is subsequently passed to the decoder of the VLM. Each decoder layer performs cross-modal attention using the standard scaled dot-product formulation, defined as \( \mathrm{Attention}(Q, K, V) = \mathrm{softmax}\left( \frac{QK^\top}{\sqrt{d_k}} \right) V \), where \( Q = H_{\mathrm{dec}} W_Q \), \( K = H_f W_K \), \( V = H_f W_V \), and \( d_k = \frac{d}{H} \), with \( H \) denoting the number of attention heads. Finally, the decoder generates a target sequence \( \hat{y} \in \mathbb{Z}^{B \times L'} \), resulting in a descriptive output \( D \) that encapsulates the idiomatic interpretation, effectively leveraging multimodal alignment and expert-guided reasoning.

\subsection{Novel Evaluation Metrics}
To systematically evaluate a model's multimodal understanding of idiomatic expressions, we introduce two metrics.

\subsubsection{Idiomatic Validation (IV) Score}
The IV Score is a structured three-stage framework that assesses (i) literal translation fidelity, (ii) visual-semantic alignment, and (iii) idiomatic meaning retention. Given an idiomatic description $d$, a pretrained VLM, specifically Qwen2.5-VL-7B, is used as a zero-shot evaluator to generate a meaning pair $\tau = \langle \xi, \mu \rangle$, where $\xi$ denotes the literal translation and $\mu$ captures its inferred meaning. The pair is sampled via $\mathcal{T} \sim \pi_{\theta'}(\tau \mid h_1, d)$, where $h_1$ is a system prompt guiding Qwen-VL's behavior. The fidelity score $\lambda$ is computed as the BLEU score between $\xi$ and the reference literal translation $\texttt{trans}(I)$, i.e., $\lambda(\xi, \texttt{trans}(I)) = \text{BLEU}(\xi, \texttt{trans}(I))$. Given the idiomatic image $\nu$, visual-semantic alignment $\delta$ with the idiom $I$ is computed as the cosine similarity between CLIP encodings of image $\phi_v(\nu)$ and idiom $\phi_t(I)$, normalized to the $[0,1]$ range: $\delta(\nu, I) = \frac{\cos(\phi_v(\nu), \phi_t(I)) + 1}{2}.$ Idiomatic meaning retention $\rho$ is quantified using BERTScore between the generated meaning $\mu$ and the ground truth: $\rho(\mu, \texttt{groundtruth}) = \text{BERTScore}(\mu, \texttt{groundtruth})$. The final IV Score is computed as the geometric mean of the three components to penalize underperformance in any one modality: $\texttt{IV\_Score}(d) = (\lambda \cdot \delta \cdot \rho)^{1/3}$. During each validation stage, Qwen2.5-VL-7B performs zero-shot inference and selects the most appropriate candidate via internal voting mechanisms, ensuring contextual fidelity and robustness in evaluation.
    
\subsubsection{\textit{IDIO-TONE} Score}
We use the evaluation dataset for generating \textit{IDIO-TONE} scores for our models (refer to Figure~\ref{fig:idio-tone-prompt} in Appendix~\ref{label:data}). The dataset is defined as $\mathcal{D} = \{(t_i, v_i, L_i)\}_{i=1}^N$, where for the $i$-th instance, $t_i$ represents the idiom, $v_i$ represents the visual input (image) and $L_i \subseteq \mathcal{C}$ is the set of ground-truth idiomatic tone labels associated with that image. The predefined label space is denoted as $\mathcal{C}$ = \{Humor, Ridicule, Affection, Aspiration, Fear, Sorrow, Deception\}. Given the idiom $t_i$ and the visual input $v_i$, our trained VLM, denoted by $f_\theta$, generates a tuple consisting of the predicted idiom translation $\hat{t}_i$ and its corresponding explanation $\hat{e}_i$, i.e., $(\hat{t}_i, \hat{e}_i) = f_\theta(t_i, v_i)$. To map the generated output back to the discrete label space $\mathcal{C}$, we employ a distilled DeepSeek-R1~\cite{guo2025deepseek} model as an evaluator, denoted by $g_\phi$, which processes the generated explanation to extract the predicted set of idiomatic labels $\hat{L}_i$, given by $\hat{L}_i = g_\phi(\hat{e}_i)$, where $\hat{L}_i \subseteq \mathcal{C}$.

To quantify the alignment between the predicted idiomatic tone labels and the gold standard, we compute a set-based $F_1$ score for each instance. For the $i$-th instance, the $F_1$ score is defined as the harmonic mean of precision and recall over the label sets. To handle the edge case where both the ground-truth and predicted sets are empty, the formulation is defined as:
\begin{equation}
    F_1^{(i)} = \begin{cases} 1, & \text{if } |L_i| = 0 \text{ \& } |\hat{L}_i| = 0,\\ \frac{2 |L_i \cap \hat{L}_i|}{|L_i| + |\hat{L}_i|}, & \text{otherwise.} \end{cases}
\end{equation}
Finally, the overall \textit{IDIO-TONE} Score, $S_{\text{IDIO-TONE}}$, for the model $f_\theta$ over the dataset $\mathcal{D}$ is calculated as the sample-averaged $F_1$-score:
\begin{equation}
    S_{\text{IDIO-TONE}} = \frac{1}{N} \sum_{i=1}^N F_1^{(i)}.
\end{equation}

\section{Experiments \& Resultant Discussion}
This section outlines the experimental setup, baseline configurations, and a head-to-head evaluation of VLMs. We supply a qualitative error analysis that pinpoints interpretive shortcomings and highlights unresolved research challenges. Our study addresses two focal research questions (RQs):
\begin{itemize}
    \item \textbf{RQ1}: To what extent does HybridMoE, in concert with the IPS module, improve idiom comprehension and contextual reasoning while preserving or boosting overall metrics?
    \item \textbf{RQ2}: What are the societal implications and cross-lingual generalizability of the proposed dataset?
\end{itemize}

The models were fine-tuned using a learning rate of $2 \times 10^{-4}$ on 3 epochs on an NVIDIA A100 GPU with 80GB VRAM. Training utilized a batch size of 4 with gradient accumulation (8 steps) to manage GPU memory efficiently. We adopted the fused AdamW optimizer to enhance generative diversity. A constant learning rate scheduler was applied for stable convergence and mixed-precision training was enabled for computational efficiency. We fine-tuned the model using PEFT with LoRA and supervised fine-tuning (SFT) to enable efficient adaptation during training and evaluation.
The dataset was split into 70\% for training, 20\% for validation, and 10\% for testing purposes. The dataset comprises 1,277 Hindi, 1,751 Thai, and 505 Bengali idiomatic samples. A comparative evaluation was performed across a range of prominent VLMs, 
including 
Blip2-7B~\cite{li2023blip2}, 
Qwen2.5-VL-7B-Instruct~\cite{qwen2.5-VL}, 
SmolVLM-Instruct~\cite{marafioti2025smolvlm},
Paligemma2-10B~\cite{paligemma2},
Llava-1.5-7B~\cite{llava1.5}, 
Gemma3-12b-pt~\cite{gemma_2025}.
Model outputs were benchmarked using a comprehensive suite of evaluation metrics designed to capture lexical overlap, semantic alignment, and distributional similarity. 
Specifically, we employed ROUGE (R1, R2, R-L, and R-Lsum) and BLEU (B1--B3 and overall score BS) to assess lexical fidelity. Semantic and structural properties were evaluated via BERTScore (BTS). Finally, we measured linguistic complexity and clarity using the IV (Idiomatic Validation) Score and the \textit{IDIO-TONE }Score.

\subsection{Results and Discussion}
This section synthesizes the findings in response to the stated RQs, supported by qualitative insights and error analyses across model generations. 
\begin{table*}[ht]
\centering
\caption{Performance variance between fine-tuned VLMs and HybridMoE-based models. Bold values denote the best scores; ↑ indicates metrics where higher is better, and ↓ denotes metrics where lower values are optimal.}
\label{tab:comp_simple_ablation}
\scalebox{.6}{
\begin{tabular}{@{}clccccccccccc@{}}
\toprule
\multirow{2}{*}{\begin{tabular}[c]{@{}c@{}}Experimental\\Setting\end{tabular}} & \multirow{2}{*}{Model Names} & \multicolumn{11}{c}{Metrics}    \\ \cmidrule(l){3-13}
 &  & \multicolumn{1}{c}{R-1} & \multicolumn{1}{c}{R-2} & \multicolumn{1}{c}{R-L} & \multicolumn{1}{c}{R-LSum} & \multicolumn{1}{c}{B-1} & \multicolumn{1}{c}{B-2} & \multicolumn{1}{c}{B-3} 
& \multicolumn{1}{c}{BS} 
& \multicolumn{1}{c}{BTS} 
& \multicolumn{1}{c}{IVS} 
& \multicolumn{1}{c}{IDIO-TONE} \\ \midrule
VLM+Finetune & Blip2-7B & 0.26 & 0.07 & 0.17 & 0.18 & 18.64 & 08.62 & 06.28 & 09.17 & 0.88 & 0.52 & 0.32 \\
 & Paligemma2-10B & 0.39 & 0.15 & 0.27 & 0.29 & 32.62 & 15.73 & 08.95 & 13.73 & 0.88 & 0.53 & 0.32 \\
 & Llava-1.5-7B & 0.35 & 0.15 & 0.26 & 0.26 & 27.46 & 12.74 & 08.20 & 11.40 & 0.90 & 0.57 & 0.31 \\
 & SmolVLM-Instruct & 0.43 & 0.18 & 0.33 & 0.33 & 47.48 & 21.82 & 14.06 & 17.70 & 0.92 & 0.59 & 0.35 \\
 & Gemma3-12b-pt & 0.4 & 0.15 & 0.29 & 0.29 & 32.86 & 13.36 & 07.90 & 11.64 & 0.90 & 0.74 & \textbf{0.54}$\uparrow$ \\
 & Qwen2.5-VL-7B-Instruct & \textbf{0.48}$\uparrow$ & \textbf{0.22}$\uparrow$ & \textbf{0.36}$\uparrow$ & \textbf{0.37}$\uparrow$ & \textbf{51.06}$\uparrow$ & \textbf{24.81}$\uparrow$ & \textbf{16.51}$\uparrow$ & \textbf{20.68}$\uparrow$ & \textbf{0.93}$\uparrow$ & \textbf{0.75}$\uparrow$ & 0.46 \\ 
 \midrule

VLM+HybridMoE & Blip2-7B & 0.3 & 0.13 & 0.23 & 0.28 & 23.76 & 12.93 & 13.29 & 15.64 & 0.89 & 0.55 & 0.33 \\
 & Paligemma2-10B & 0.45 & 0.2 & 0.34 & 0.35 & 35.97 & 17.25 & 14.2 & 15.94 & 0.89 & 0.59 & 0.32 \\
 & Llava-1.5-7B & 0.4 & 0.17 & 0.32 & 0.34 & 32.52 & 16.64 & 12.95 & 14.25 & 0.91 & 0.61 & 0.33 \\
 & SmolVLM-Instruct & 0.48 & 0.21 & 0.36 & 0.36 & 50.46 & 24.16 & 16.59 & 20.92 & 0.92 & 0.66 & 0.35 \\
 & Gemma3-12b-pt & 0.46 & 0.19 & 0.35 & 0.35 & 38.81 & 17.82 & 11.93 & 14.58 & 0.92 & 0.79 & \textbf{0.55}$\uparrow$ \\
 & Qwen2.5-VL-7B-Instruct & \textbf{0.54}$\uparrow$ & \textbf{0.32}$\uparrow$ & \textbf{0.41}$\uparrow$ & \textbf{0.43}$\uparrow$ & \textbf{56.32}$\uparrow$ & \textbf{28.92}$\uparrow$ & \textbf{21.48}$\uparrow$ & \textbf{24.19}$\uparrow$ & \textbf{0.95}$\uparrow$ & \textbf{0.82}$\uparrow$ & 0.47 \\ 
\bottomrule
\end{tabular}%
}
\end{table*}

\begin{figure*}[t]    
\centering
  \includegraphics[width=\textwidth]{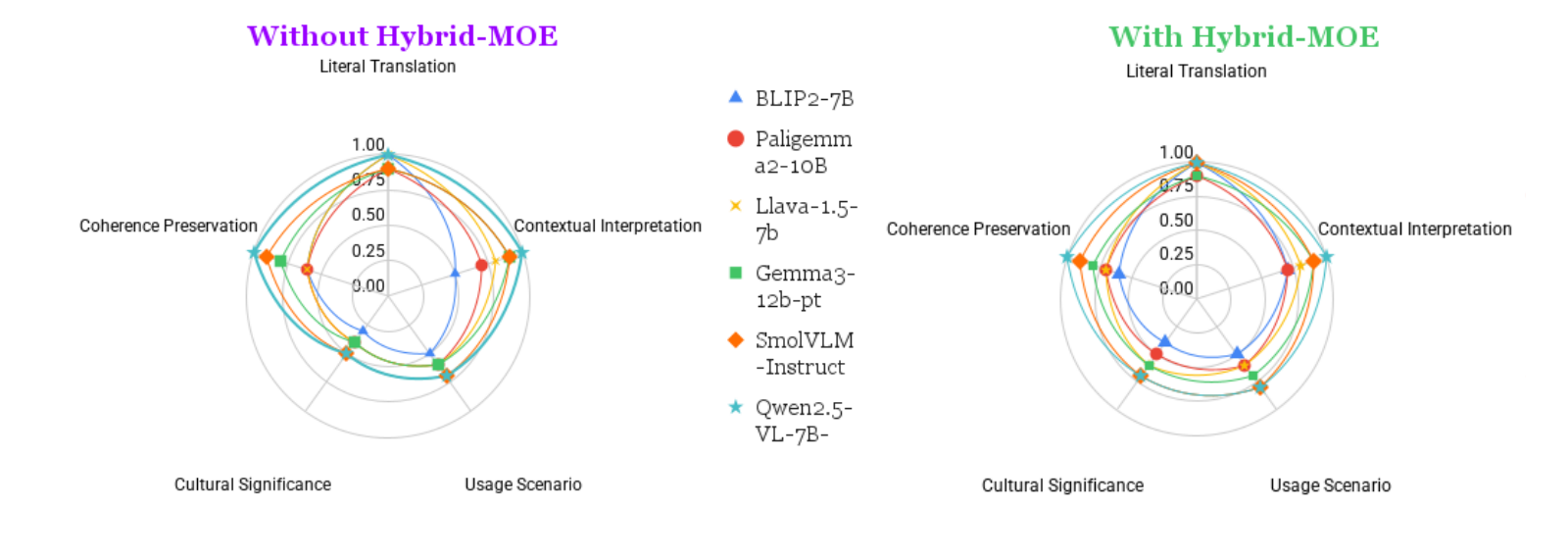}
    \caption{Expert Evaluation of Idiom Understanding Performance of VLM models with Idiomatic Property Retention (IPR) Criteria.}
    \label{win}
\end{figure*}
\subsubsection{Response to RQ1--Impact of HybridMoE}
The ablation results in Table~\ref{tab:comp_simple_ablation} reveal a marked elevation in idiomatic reasoning performance when transitioning from conventional fine-tuning to the proposed HybridMoE framework. Standard VLMs exhibit moderate gains in syntactic fidelity and surface-level lexical overlap; however, their ability to generalize over complex, culturally embedded figurative constructs remains constrained, particularly in low-resource multilingual contexts. By contrast, HybridMoE introduces a dynamic expert routing topology that enables token-level modulation across modality- and task-specialized subspaces, yielding significantly sharper representations for idiomatic abstraction and cultural disambiguation. Qwen2.5-VL-7B-Instruct + HybridMoE exhibits significant performance gains. 
ROUGE-L improves from 0.36 to 0.41, BLEU-3 from 16.51 to 21.48, and  
BERTScore from 0.93 to 0.95,
indicating enhanced semantic precision and idiomatic fluency. Qwen2.5-VL-7B+HybridMoE has the highest IV (Idiomatic Validation) score of 0.82. Qualitative analysis further reveals improved figurative anchoring within culturally grounded narratives. Similar gains in SmolVLM-Instruct and Gemma3-12b-pt confirm that HybridMoE's modular design offers architecture-agnostic improvements by enabling cross-modal grounding, idiom-specific alignment, and reducing overfitting through shared expert knowledge. However, the \textit{IDIO-TONE} results reveal a clear limitation in modeling idiomatic tone, with most models scoring around $\sim$0.30-0.35. Even the best-performing models, such as \textit{Gemma3-12b-pt} (0.55) and \textit{Qwen2.5-VL-7B-Instruct} (0.47), achieve only moderate improvements. This highlights a persistent gap in capturing pragmatic and culturally grounded nuances, indicating the need for more targeted learning strategies. All reported results are statistically significant at $p<0.05$~\cite{welch1947generalization}. 

Notably, the efficacy of the HybridMoE architecture, instantiated with Qwen2.5-VL-7B-Instruct, is further substantiated through evaluation on the multimodal V-FLUTE~\cite{saakyan2024v} corpus. As shown in Table~\ref{tab:sft_comparison}, HybridMoE consistently yields significant performance gains over standard supervised fine-tuning (SFT).

\subsubsection{Response to RQ2--Generalizability and Societal Relevance}
The \textit{Varnika} corpus serves as a comprehensive multimodal benchmark for idiomatic understanding across Hindi, Thai, and Bengali. As evidenced in Table~\ref{tab:comp_simple_ablation}, even VLMs not explicitly designed for multilingual settings demonstrate strong adaptability to \textit{Varnika}, achieving effective idiomatic interpretation under both fine-tuning and HybridMoE configurations. Furthermore, as shown in Table~\ref{tab:sft_comparison}, language-specific and relatively low-resource models such as \textit{Ganga} (Hindi)~\cite{lingo-research-group-at-iit-gandhinagar-india} and \textit{Typhoon} (Thai)~\cite{pipatanakul2023typhoon} also exhibit the ability to infer idiomatic meanings when trained on \textit{Varnika}. However, \textit{TitulLaMA} (Bengali)~\cite{nahin2025titullmsfamilybanglallms} underperforms, likely due to its pretraining bias toward Bengali combined with the relatively lower representation of Bengali idioms in the dataset. Notably, despite their smaller scale, these language-specific models achieve \textit{IDIO-TONE} scores comparable to larger LLMs and VLMs (Table~\ref{tab:comp_simple_ablation}), suggesting that idiomatic tone understanding is not solely dependent on model size but also on alignment with linguistic and cultural context. These findings demonstrate the strong generalizability of \textit{Varnika} across diverse model architectures and linguistic settings. Beyond technical performance, \textit{Varnika} holds substantial educational and societal value.

\begin{table*}[ht]
\centering
\caption{Additional Experiments on language-specific LLMs and also SFT on Flute and V-Flute dataset.}
\label{tab:sft_comparison}
\scalebox{0.6}{
\begin{tabular}{@{}llccccccccccc@{}}
\toprule
\multirow{2}{*}{\begin{tabular}[c]{@{}c@{}}Experimental\\ Setting\end{tabular}} & \multirow{2}{*}{Model Names} & \multicolumn{11}{c}{Metrics} \\ \cmidrule(l){3-13} 
 &  & R-1 & R-2 & R-L & R-LSum & B-1 & B-2 & B-3 & BS & BTS & IVS & IDIO-TONE \\ \midrule
\multirow{3}{*}{SFT on \textit{Varnika}} & Ganga-2-1B & 0.314 & 0.127 & 0.268 & 0.269 & 33.56 & 13.13 & 7.83 & 12.84 & 0.83 & 0.32 & 0.31 \\
 & Titulm-Llama-3.2-1B& 0.025 & 0.01 & 0.018 & 0.02 & 2.13 & 0.16 & 0.031 & 0.15 & 0.8 & 0.25 & 0.28 \\
 & Typhoon-7B-Instruct & 0.293 & 0.18 & 0.248 & 0.249 & 19.32 & 12.32 & 9.53 & 12.11 & 0.88 & 0.51 & 0.46 \\ \midrule
SFT on Flute & Qwen2.5-VL-7B-Instruct & 0.39 & 0.21 & 0.33 & 0.33 & 27.43 & 16.77 & 11.92 & 18.41 & 0.84 & 0.58 & - \\ \midrule
SFT on V-flute & Qwen2.5-VL-7B-Instruct & 0.34 & 0.19 & 0.22 & 0.26 & 21.14 & 11.82 & 9.02 & 15.59 & 0.75 & 0.44 & - \\
\midrule 
HybridMoE on V-flute & Qwen2.5-VL-7B-Instruct & 0.43 & 0.21& 0.35 & 0.35& 46.59& 18.13 & 9.80 & 14.34 & 0.89 & 0.59 & - \\
\bottomrule
\end{tabular}
}
\end{table*}

\subsection{Analytical Discussion}

\begin{figure}[t]
    \centering
    \begin{subfigure}[t]{0.48\columnwidth}
        \centering
        \includegraphics[width=0.8\textwidth]{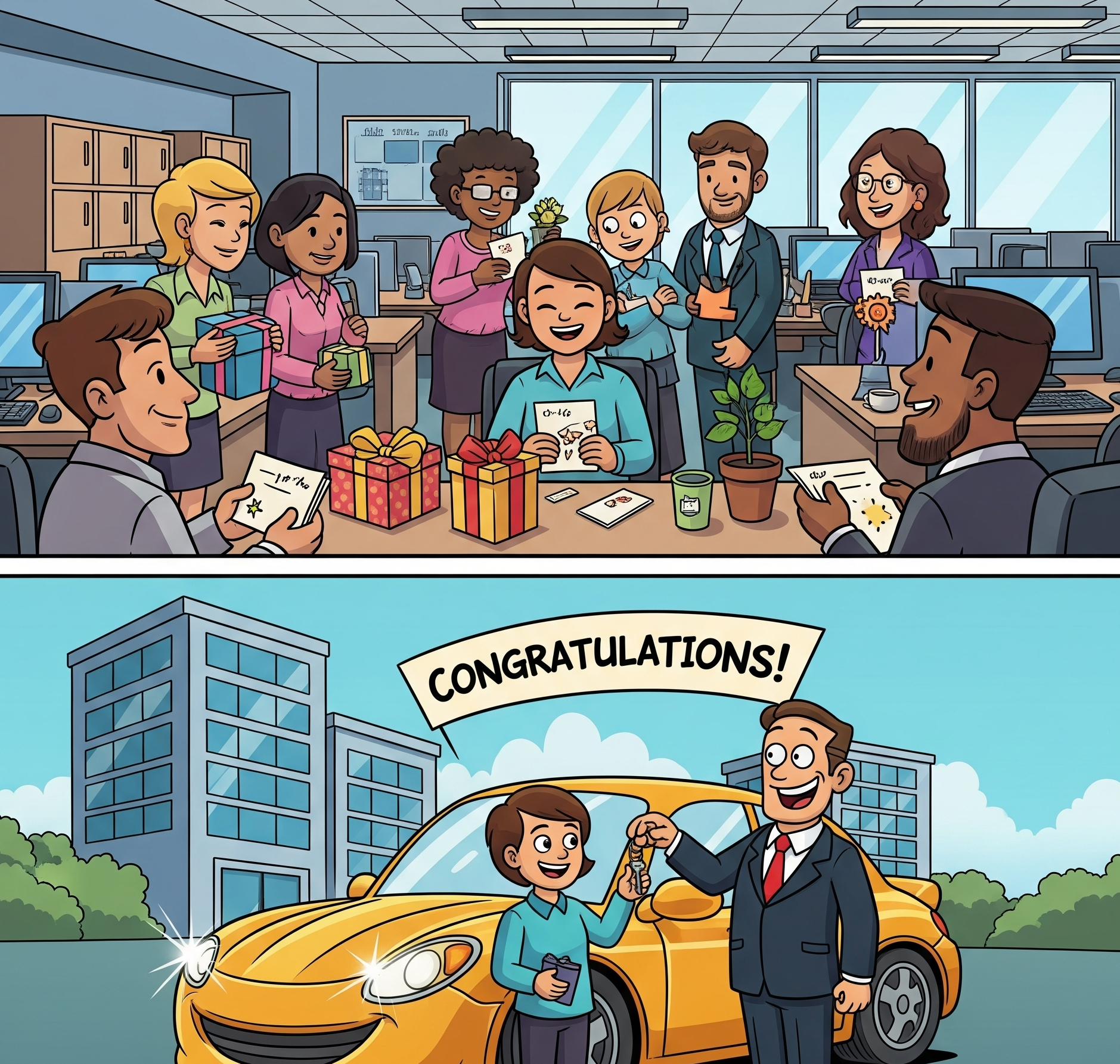}
        \caption{Qualitative Analysis of the Hindi idiom 
        \begin{hindi}सोने पे सुहागा\end{hindi} 
        (\textit{Sone pe Suhāgā}; literal meaning: gold with a touch of \textit{suhāgā}, a traditional polishing substance; metaphorical meaning: an added bonus or improvement to something already good)}
        \label{fig:qualitative}
    \end{subfigure}
    \hfill
    \begin{subfigure}[t]{0.48\columnwidth}
        \centering
        \includegraphics[width=0.8\textwidth]{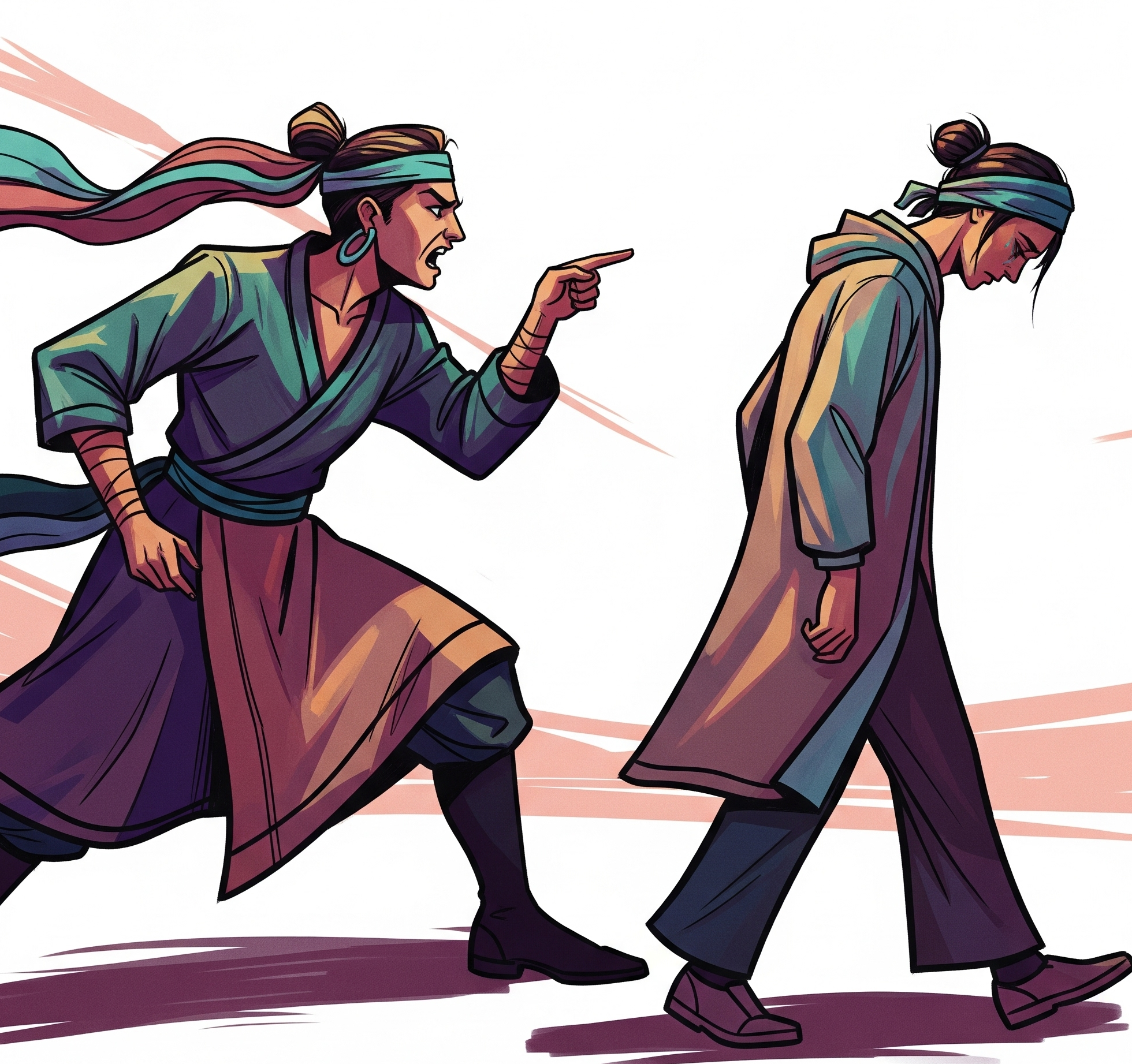}
        \caption{Error Analysis of the Hindi idiom 
        \begin{hindi}दफा होना\end{hindi} 
        (\textit{Dafā honā}; literal meaning: to disappear or go away; metaphorical meaning: dismissal or rejection, implying exclusion)}
        \label{fig:error}
    \end{subfigure}
    \caption{Comparative Qualitative and Error Analyses of Idiom Understanding in Vision Language Models}
    \label{fig:qual_error_combined}
\end{figure}

\subsubsection{Human Evaluation} 
We conducted a human evaluation involving three native-speaking cultural experts with domain expertise in Hindi, Bengali, and Thai, who assessed 709 instances using Information Persistence Ratings (IPR)~\cite{das2026meaningisntliteralexploring} across five dimensions: literal accuracy, contextual fit, usage naturalness, cultural depth, and overall coherence. As depicted in Figure~\ref{win}, models enhanced with HybridMoE consistently outperformed their finetuned counterparts. Qwen2.5-VL-7B-Instruct + HybridMoE showed the most significant gains, particularly in contextual interpretation (+0.1), cultural significance (+0.2), and usage scenario (+0.1). These improvements stem from HybridMoE’s dynamic expert routing and task-specific reasoning, enabling better disentanglement of semantic, cultural, and contextual representations. In contrast, conventional fine-tuning lacks such specialization, leading to shallow generalizations. Consistent patterns observed in SmolVLM-Instruct and LLaVA-1.5-7B further substantiate HybridMoE as a resilient framework for cross-modal idiomatic reasoning in low-resource, culturally nuanced settings.

\subsubsection{Qualitative Analysis}
Figure~\ref{fig:qualitative} highlights that models integrated with HybridMoE exhibit significantly improved idiomatic understanding. While finetuned models tend to rely on object-centric cues (e.g., gold, gems) and fail to capture the relational enhancement (good → better), HybridMoE-augmented models perform stronger semantic alignment and figurative abstraction. Interestingly, Qwen2.5-VL-7B-Instruct + HybridMoE, SmolVLM-Instruct + HybridMoE, and Gemma3-12b-pt + HybridMoE accurately capture both the literal origin and idiomatic meaning, often producing coherent visual metaphors of transformation or enhancement. These improvements stem from HybridMoE’s dynamic expert routing, which activates culturally informed reasoning pathways for precise and context-aware idiom interpretation.

\subsubsection{Error Analysis}
Figure~\ref{fig:error} illustrates a failure case where current VLMs struggle to grasp the idiomatic meaning of \begin{hindi}दफा होना\end{hindi}. These models tend to rely on shallow lexical or visual cues, produce interpretations that miss the cultural and contextual depth of the idiom. For example, SmolVLM-Instruct and LLaVA-1.5-7B generated responses such as ``to be on the right path'' or ``to go through a cycle,'' which are far from the idiom's actual meaning of disappearance or being dismissed. These mismatches highlight the shortcomings of standard fine-tuning when it comes to handling figurative, culturally grounded language. In contrast, Qwen2.5-VL-7B-Instruct enhanced with HybridMoE effectively captures both the literal and figurative aspects, producing a meaningful visual metaphor, such as pushing away shadowy figures that accurately reflect the idiom's true intent.

\section{Conclusion}
This paper introduces a HybridMoE architecture that enhances idiomatic understanding with IPS in VLMs through task-adaptive expert routing. Alongside this framework, we present \textit{Varnika} \begin{hindi}(वर्णिका)\end{hindi}, a reconstructed extension of the multimodal idiom dataset \textit{Mediom}~\cite{das2026meaningisntliteralexploring}, enriched with fine-grained annotations across seven idiomatic tones such as \textit{Humor}, \textit{Ridicule}, \textit{Affection}, \textit{Aspiration}, \textit{Fear}, \textit{Sorrow}, and \textit{Deception}.  Empirical results show improved lexical abstraction, figurative disambiguation, and cultural grounding in models, such as Qwen2.5-VL-7B, validated by the proposed IV (Idiomatic Validation) and \textit{IDIO-TONE} scores. \textit{Varnika} advances inclusive, culturally aware AI while enabling applications in creative modeling and education for social good.

\section{Limitations}
Despite the improvements introduced by the HybridMoE framework and the \textit{Varnika} dataset, several limitations remain:

\begin{itemize}
    \item \textbf{Dependence on Visual Grounding:} The idiomatic reasoning capabilities of current VLMs remain sensitive to the availability of meaningful visual cues. For highly abstract or culturally implicit idioms lacking clear visual metaphors, models often default to surface-level or literal interpretations. This exposes a limitation of HybridMoE, which, despite enhanced feature routing, does not fully bridge deeper figurative reasoning gaps.

    \item \textbf{Subjectivity in IDIO-TONE Annotations:} The proposed seven-tone taxonomy captures broad pragmatic dimensions; however, idiomatic tone interpretation is inherently subjective. Fine-grained emotional nuances and hierarchical relationships between tones are not explicitly modeled, which may limit the expressiveness and consistency of the \textit{IDIO-TONE} score.

    \item \textbf{Limitations of IV (Idiomatic Validation) Score:} The Idiomatic Validation (IV) score, while effective in combining literal fidelity, semantic alignment, and multimodal grounding, is an aggregated metric. It may obscure component-level weaknesses, thereby reducing the interpretability of model-specific strengths and failure modes.

    \item \textbf{Static Evaluation Setting:} The dataset and evaluation framework are static, whereas idiomatic meaning is dynamic and context-dependent. This limits the assessment of model performance in real-world conversational or evolving discourse scenarios.
\end{itemize}

While HybridMoE enhances idiomatic abstraction through expert specialization, it does not fully bridge the gap between cultural reasoning and perceptual grounding. The results indicate that idiomatic understanding remains challenging for VLMs without explicit cultural or visual cues, highlighting the need for integrating external knowledge and advanced reasoning mechanisms beyond static image-text alignment.

\section{Ethical Considerations}
While \textit{Varnika} enhances the \textit{Mediom} dataset through the integration of fine-grained idiomatic tonal annotations, several ethical considerations arise from the process of tone-centric labeling:

\begin{enumerate}
    \item \textbf{Subjectivity in Tonal Interpretation:} Idiomatic tone assignment inherently involves subjective judgment, as tones such as \textit{Humor}, \textit{Ridicule}, or \textit{Affection} may be perceived differently across individuals and cultural contexts. Despite expert validation and annotation guidelines, residual interpretative variability may influence label consistency.

    \item \textbf{Cultural Sensitivity and Misrepresentation:} Idioms are deeply embedded in cultural, historical, and social contexts. Assigning discrete tonal categories risks oversimplifying or misrepresenting nuanced cultural meanings, particularly for expressions that carry layered or context-dependent interpretations across Hindi, Bengali, and Thai.

    \item \textbf{Multi-Label Ambiguity and Overlap:} Many idioms naturally express multiple overlapping tones (e.g., \textit{Ridicule} intertwined with \textit{Humor} or \textit{Sorrow}). While the multi-label framework captures this complexity, it may still fail to fully represent subtle tonal gradations or hierarchical relationships between tones.

    \item \textbf{Bias in Tonal Distribution:} The frequency and distribution of selected \textit{IDIO-TONE} categories may reflect underlying biases in data sources or annotation practices, potentially leading to over-representation of certain emotional or pragmatic tones while under-representing others.

    \item \textbf{Static Representation of Dynamic Semantics:} Idiomatic tones are context-sensitive and may evolve across discourse, media, and cultural settings. The current dataset provides a static snapshot of tonal interpretations, which may not fully capture the dynamic and evolving nature of idiomatic usage in real-world communication.
\end{enumerate}

\section{Future Work}
In future work, we aim to extend \textit{Varnika} by expanding its coverage to a broader range of Asian languages, enabling richer cross-lingual and cultural analysis of idiomatic understanding. We also plan to explore reinforcement learning (RL)-based frameworks to model idiomatic reasoning as a dynamic, feedback-driven process within multimodal settings. Additionally, we seek to enhance idiomatic interpretation by incorporating more refined semantic and pragmatic modeling, including context-aware reasoning and structured tone representations.

\section{Acknowledgement}
All the authors sincerely thank the dataset annotators Tannu, Jheel, and Paavnee for their valuable contributions in curating and validating the idiomatic tone annotations. Their efforts were instrumental in ensuring the quality and reliability of the \textit{Varnika} dataset.

\bibliography{custom}

\appendix

\section{Appendix}
\label{sec:appendix}

Idioms constitute a rich form of figurative expression, wherein meaning extends beyond literal composition to encode culturally grounded emotions, social intent, and contextual nuance. Their interpretation is inherently tied not only to semantics but also to underlying affective and pragmatic signals, making tone an essential dimension of idiomatic understanding. For instance, the Hindi idiom \begin{hindi}ऊँट के मुँह में जीरा\end{hindi} (\textit{oont ke muh mein jeera}, ``a cumin seed in a camel’s mouth'') conveys insufficiency and is often associated with tones of \textit{Ridicule} and subtle \textit{Sorrow}, reflecting dissatisfaction in disproportionate situations. Similarly, the Bengali idiom \begin{bengali}নাচতে না জানলে উঠোন বাঁকা\end{bengali} (\textit{nachte na janle uthoon banka}, ``blaming the courtyard for not knowing how to dance'') embodies \textit{Ridicule} and \textit{Deception}, exposing the human tendency to deflect personal shortcomings. 

In contrast, idioms may also encode positive or aspirational sentiments. The Thai idiom \begin{thai}ช้า ๆ ได้พร้าเล่มงาม\end{thai} (\textit{cha cha dai phra lem ngam}, ``slowly, one obtains a beautiful blade'') reflects patience and perseverance, aligning with tones of \textit{Aspiration} and calm \textit{Affection} toward disciplined effort. Likewise, expressions such as \textit{breaking the ice} often carry \textit{Humor} and social \textit{Affection}, facilitating interpersonal connection. Conversely, idioms like \textit{cry over spilled milk} encapsulate \textit{Sorrow}, while \textit{a wolf in sheep’s clothing} signals \textit{Deception} and latent \textit{Fear}, highlighting cautionary undertones embedded in everyday language.

These examples illustrate that idioms inherently operate across a spectrum of emotional and pragmatic tones, including \{\textit{Humor}, \textit{Ridicule}, \textit{Affection}, \textit{Aspiration}, \textit{Fear}, \textit{Sorrow}, \textit{Deception}\} which shape their intended meaning and usage. Therefore, capturing idiomatic meaning without accounting for such tonal dimensions results in incomplete or misleading interpretations. Despite this, existing research remains predominantly English-centric~\cite{haagsma2020magpie}, with limited focus on multilingual and multimodal settings where emotional, cultural, and visual grounding jointly influence idiomatic understanding.

\subsection{Justification of the selected languages}

The selection of Hindi, Bengali, and Thai in this work is motivated by a unique combination of typological diversity and deep-rooted cultural interconnectedness across South and Southeast Asia. While Hindi and Bengali belong to the Indo-Aryan family and Thai is a tonal, analytic language from the Kra-Dai family, these languages are historically linked through centuries of cultural exchange shaped by Sanskrit and Pali traditions. This shared heritage is reflected not only in vocabulary but also in religious narratives, philosophical doctrines, and storytelling traditions. For instance, epics such as the \textit{Ramayana} are preserved in India and reinterpreted in Thailand as the \textit{Ramakien}, demonstrating parallel mythological structures and moral frameworks that influence idiomatic expressions and figurative language.

Such cultural alignment manifests in idioms that encode similar emotional and pragmatic intents despite linguistic variation. Expressions conveying moral causality, fate, or human behavior often draw upon shared belief systems such as karma, duty, and social conduct. These idioms are not merely linguistic constructs but carriers of culturally embedded sentiments ranging from cautionary \textit{Fear} and reflective \textit{Sorrow} to social \textit{Ridicule}, aspirational values, and interpersonal \textit{Affection}. 

This cross-cultural consistency directly motivates the adoption of the seven idiomatic tone categories \{\textit{Humor}, \textit{Ridicule}, \textit{Affection}, \textit{Aspiration}, \textit{Fear}, \textit{Sorrow}, \textit{Deception}\} as they capture the most recurrent and semantically stable pragmatic dimensions observed across these languages. Alternative tonal categories were explored during dataset construction; however, they either exhibited significant overlap or lacked consistent representation across linguistic and cultural contexts. In contrast, the selected seven tones provide a balanced and expressive framework for modeling idiomatic meaning, aligning closely with the emotional and narrative functions embedded in these traditions.

From a modeling perspective, incorporating Thai introduces a controlled typological contrast, enabling the evaluation of whether models can generalize idiomatic reasoning beyond closely related language families. Success in transferring idiomatic understanding across these languages indicates abstraction beyond lexical patterns toward deeper cultural and semantic comprehension. Therefore, the inclusion of Hindi, Bengali, and Thai not only enhances linguistic diversity but also establishes a culturally grounded and theoretically meaningful testbed for studying multimodal idiomatic reasoning.

\subsection{Dataset}\label{label:data}
Building upon the existing \textit{Mediom} dataset, which already captures rich syntactic diversity across idioms in Hindi, Bengali, and Thai, we extend it by incorporating fine-grained pragmatic tonality annotations. The dataset includes a wide range of idiomatic constructions, such as fixed expressions (e.g., \begin{thai}น้ำท่วมปาก\end{thai}, \textit{unable to speak out}), semi-fixed constructions (e.g., \begin{bengali}নিজের পায়ে কুড়াল মারা\end{bengali}, \textit{to harm oneself}), verb–object structures (e.g., \begin{hindi}नाक रगड़ना\end{hindi}, \textit{to plead intensely}), adjective–noun phrases (e.g., \begin{bengali}মিষ্টি স্বপ্ন\end{bengali}, \textit{sweet dreams}), prepositional idioms (e.g., \begin{thai}เข้าหูซ้ายทะลุหูขวา\end{thai}, \textit{in one ear and out the other}), and binomial expressions (e.g., \begin{hindi}उल्टा सीधा\end{hindi}, \textit{nonsensical behavior}).

While \textit{Mediom} effectively captures structural and semantic richness, it does not explicitly model the underlying emotional and pragmatic tones that govern idiomatic usage. To address this gap, we augment each instance with a set of idiomatic tone labels, grounded in a predefined taxonomy. During the initial design phase, we explored a broader set of ten tonal categories, including additional dimensions such as \textit{Surprise}, \textit{Anger}, and \textit{Neutral}. However, upon systematic analysis and expert-driven validation, these categories were found to be either semantically overlapping with existing tones or inconsistently represented across languages and modalities. 

\begin{figure*}[!t]
\centering
\begin{tikzpicture}
\begin{axis}[
    ybar,
    bar width=12pt,
    width=0.95\textwidth,
    height=8cm,
    ymin=0,
    ymax=25,
    ylabel={Percentage (\%)},
    xlabel={Categories},
    xlabel style={font=\bfseries},
    ylabel style={font=\bfseries},
    title style={font=\bfseries, yshift=2pt},
    symbolic x coords={Humorous, Ridicule, Affection, Aspiration, Fear, Sorrow, Deception},
    xtick=data,
    xticklabel style={font=\normalsize},
    ytick distance=5,
    enlarge x limits=0.05,
    enlarge x limits=0.15, 
    area legend,    
    legend style={
        at={(0.98,0.98)},
        anchor=north east,
        draw=gray!30,
        fill=white,
        font=\small
    },
    nodes near coords,
    every node near coord/.append style={
        font=\tiny,
        yshift=2pt,
        text=black
    },
    point meta=explicit symbolic,
    axis line style={black!60},
    tick style={black!60},
    ymajorgrids=true,
    xmajorgrids=false,
    grid style=dashed,
]

\addplot[
    fill=red!35,
    draw=black!60
] coordinates {
    (Humorous,8.40) [8.4]
    (Ridicule,11.40) [11.4]
    (Affection,11.00) [11.0]
    (Aspiration,17.83) [17.8]
    (Fear,19.34) [19.3]
    (Sorrow,21.00) [21.0]
    (Deception,11.03) [11.0]
};
\addlegendentry{Hindi}

\addplot[
    fill=blue!45,
    draw=black!60
] coordinates {
    (Humorous,7.20) [7.2]
    (Ridicule,10.30) [10.3]
    (Affection,9.40) [9.4]
    (Aspiration,22.00) [22.0]
    (Fear,20.54) [20.5]
    (Sorrow,17.70) [17.7]
    (Deception,12.86) [12.9]
};
\addlegendentry{Bengali}

\addplot[
    fill=green!45,
    draw=black!60
] coordinates {
    (Humorous,8.55) [8.6]
    (Ridicule,11.53) [11.5]
    (Affection,11.82) [11.8]
    (Aspiration,18.51) [18.5]
    (Fear,19.34) [19.3]
    (Sorrow,17.70) [17.7]
    (Deception,12.55) [12.6]
};
\addlegendentry{Thai}

\end{axis}
\end{tikzpicture}
\caption{Distribution of Idiomatic Tone categories across Hindi (h), Bengali (b), and Thai (t).}
\label{fig:idiotone_distribution}
\end{figure*}
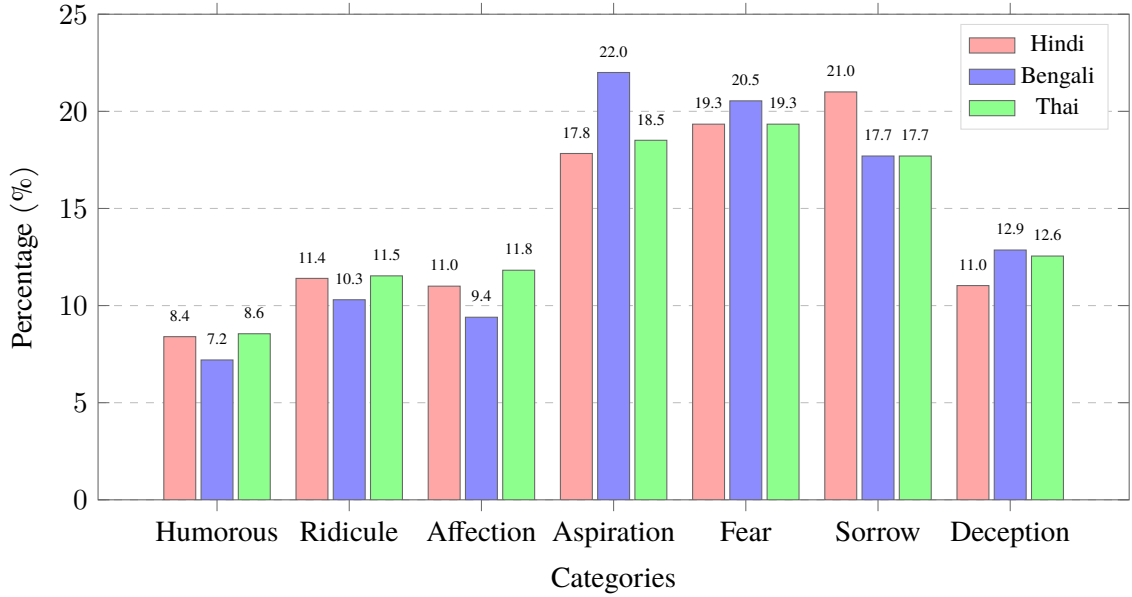

Consequently, we refine the taxonomy to seven core tones \{\textit{Humor}, \textit{Ridicule}, \textit{Affection}, \textit{Aspiration}, \textit{Fear}, \textit{Sorrow}, \textit{Deception}\}, as depicted in Figure~\ref{fig:idiotone_distribution} and demonstrate higher cross-lingual consistency, clearer semantic boundaries, and stronger alignment with both textual and visual representations. This augmentation enables a more comprehensive representation of idiomatic meaning by integrating structural, semantic, and affective dimensions within a unified multimodal framework.

\subsubsection{Data Quality Assurance}\label{dataset_quality}
To ensure high-quality annotations and robust validation of cross-modal tonal alignment, we engaged a team of domain experts comprising one doctoral researcher and two literature professors, all of whom are native speakers of Thai, Hindi, and Bengali. Their combined linguistic proficiency and academic expertise enabled the precise mapping of culturally nuanced idiomatic expressions to their corresponding pragmatic tone labels.

\subsubsection{Annotation Procedure} 
The annotation process was designed to systematically capture the pragmatic and emotional tones of idioms in the \textit{Mediom} dataset. Each annotator was provided with the complete multimodal instance, including the idiom text, its literal translation, figurative meaning, idiomatic interpretation, and the corresponding generated image. The goal was to assign one or more labels from a predefined idiomatic tone taxonomy:
$\mathcal{C}$ = \{Humor, Ridicule, Affection, Aspiration, Fear, Sorrow, Deception\}

To establish a reliable foundation, expert annotators first independently labeled a subset of 100 samples. Each instance was carefully examined using its full multimodal context, and multiple tone labels were assigned when necessary. This process resulted in a high-quality reference set of 300 gold-standard annotations, which was subsequently used to guide the broader annotation effort.

To ensure annotation quality and consistency, we developed a structured evaluation framework focusing on five key aspects:
\begin{enumerate}[label=(\roman*)]
    \item Accurate understanding of tone from multimodal inputs.
    \item Consistency of tone across text and image.
    \item Alignment with real-world usage and intended meaning.
    \item Preservation of cultural and pragmatic nuances.
    \item Coherence among multiple assigned tones.
\end{enumerate}
This framework served as a guiding principle for both annotation and validation.

Following this, a group of trained undergraduate annotators conducted the main annotation process. They were first introduced to a curated set of reference examples and detailed annotation guidelines, enabling them to understand how to interpret and assign idiomatic tones effectively. During this phase, annotators received continuous feedback and clarification from the expert panel to resolve ambiguities and maintain consistency.

Finally, all annotated instances underwent a rigorous validation stage. Expert annotators reviewed and verified each annotation to ensure semantic correctness, cultural fidelity, and adherence to the defined \textit{IDIO-TONE} taxonomy. This two-stage pipeline annotation followed by expert validation ensured high-quality, reliable, and culturally grounded labels across the dataset. Annotators were compensated at a rate of \$0.5 per sample.
\begin{enumerate}[label=(\roman*)]
\item \textbf{Peer Review:} Each annotated instance was independently reviewed by at least two additional fluent annotators to minimize subjective bias and ensure consistency in tone assignment. Furthermore, to evaluate cross-modal pragmatic alignment, we quantified the overlap between tone labels assigned based on textual interpretations and those inferred from corresponding visual representations as described in Figure~\ref{fig:idio-tone-prompt}. This analysis yielded an overlap of \textbf{54.75\%}, indicating a strong correlation between idiomatic tonal intent and its visual grounding.

\begin{figure*}[ht]
\begin{promptbox}{\textbf{Validation Centric Idiomatic Tonality Label Generation from Textual Idioms}}
\textbf{System Role:} You are a linguistic expert and expert at multi-cultural idiomatic understanding.

\textbf{Instruction:} You will be provided with an idiom, along with its English translation and meaning. It is your responsibility to figure out which of the following labels best fits the given idiom. Here are the labels: Humorous, Ridicule, Affection, Aspiration, Fear, Sorrow, Deception
Here is the output format: 
{label0: reasoning for assigning label0,
 label1: reasoning for assigning label1,
\ldots}

Return nothing, except the required dictionary.
\end{promptbox}

\begin{promptbox}{\textbf{Validation Centric Idiomatic Tonality Label Generation from Images}}
\textbf{System Role:} You are an expert image analyzer and cross-cultural idiomatic expert.

\textbf{Instruction:} You will be provided with an image which represents an idiom, along with its English translation and meaning. It is your responsibility to figure out which of the following labels best fits the given idiom. Here are the labels: Humorous, Ridicule, Affection, Aspiration, Fear, Sorrow, Deception.
Here is the output format:
{label0: reasoning for assigning label0,
label1: reasoning for assigning label1,
\ldots}

Return nothing, except the required dictionary.

\textbf{Input:} \{Output of Stage 1\}
\end{promptbox}
\caption{Idiomatic Tonality label validations by VLM DeepSeek-R1}\label{fig:idio-tone-prompt}
\end{figure*}
\item \textbf{Expert Validation:} A panel of cultural and linguistic experts conducted a comprehensive final review of the annotated samples. Each instance was evaluated against the predefined assessment criteria, with scores assigned based on the extent to which the five aspects of tonal alignment were preserved (e.g., a score of 5 indicates full compliance across all criteria).
\end{enumerate}

The final annotation process achieved a substantial inter-annotator agreement score of 0.65 (Cohen’s kappa~\cite{gwet2014handbook}), reflecting a high degree of consistency and reliability in the tone annotation framework.  

\subsection{Results \& Analysis}
To understand the precise contribution of the IPS in our HybridMoE framework, we conduct a rigorous ablation comparing two configurations: (i) HybridMoE-augmented VLMs without IPS, and (ii) full HybridMoE with IPS injection in Table~\ref{tab:comp_Ips}. The comparative results indicate that HybridMoE-augmented VLMs, even without IPS injection, provide a strong baseline improvement across models by enhancing multimodal fusion and representation learning. In this setting, models such as \textit{Qwen2.5-VL-7B-Instruct} and \textit{Gemma3-12B-pt} achieve competitive performance on lexical and sequence-level metrics, with ROUGE-LSum values around 0.41--0.43 and BLEU-3 scores exceeding 21, reflecting improved coherence and n-gram alignment. However, despite these gains, the absence of IPS limits the model’s ability to capture deeper semantic and idiomatic nuances. This is evident from comparatively moderate BERTScore values ($\approx 0.89$--$0.95$) and lower Idiomatic Validation Scores and \textit{IDIO-TONE} metrics, suggesting that the improvements are largely driven by surface-level alignment rather than true figurative understanding.

\begin{table*}[!t]
\centering
\caption{The variance in performance metrics between  HybridMoE-enhanced and HybridMoE-without IPS is presented below. Here, R-1, R-2, R-L, and R-LSum denote ROUGE-1, ROUGE-2, ROUGE-L, and ROUGE-LSum scores, respectively. B-1, B-2, B-3, and BS denote BLEU-1, BLEU-2, BLEU-3, and the BLEU score. BTS, IVS, and \textit{IDIO-TONE} correspond to BERTScore, Idiomatic Validation Score, and the proposed idiomatic tone score. Bold values indicate the best performance; $\uparrow$ denotes that higher values are better.}
\label{tab:comp_Ips}
\scalebox{0.6}{
\begin{tabular}{@{}clccccccccccc@{}}
\toprule
\multirow{2}{*}{\begin{tabular}[c]{@{}c@{}}Experimental\\Setting\end{tabular}} 
& \multirow{2}{*}{Model Names} 
& \multicolumn{11}{c}{Metrics} \\ 
\cmidrule(l){3-13}
&  & R-1 & R-2 & R-L & R-LSum & B-1 & B-2 & B-3 & BS & BTS & IVS & IDIO-TONE \\ 
\midrule
\multirow{6}{*}{\begin{tabular}[c]{@{}c@{}}HybridMoE\\w/o IPS\end{tabular}}
& Blip2-7B & 0.27 & 0.10 & 0.19 & 0.23 & 19.91 & 9.62 & 8.02 & 11.73 & 0.88 & 0.53 & 0.33 \\
& Paligemma2-10B & 0.42 & 0.17 & 0.30 & 0.30 & 33.20 & 16.64 & 11.82 & 14.80 & 0.89 & 0.56 & 0.32 \\
& Llava-1.5-7B & 0.37 & 0.15 & 0.27 & 0.29 & 29.53 & 14.91 & 9.36 & 12.94 & 0.90 & 0.58 & 0.33 \\
& SmolVLM-Instruct & 0.44 & 0.18 & 0.35 & 0.35 & 48.92 & 22.75 & 15.37 & 19.00 & 0.92 & 0.63 & 0.36 \\
& Gemma3-12b-pt & 0.42 & 0.17 & 0.31 & 0.32 & 35.04 & 15.43 & 8.99 & 12.69 & 0.91 & 0.76 & \textbf{0.54}$\uparrow$ \\
& Qwen2.5-VL-7B-Instruct & \textbf{0.49}$\uparrow$ & \textbf{0.23}$\uparrow$ & \textbf{0.39}$\uparrow$ & \textbf{0.40}$\uparrow$ & \textbf{53.22}$\uparrow$ & \textbf{25.90}$\uparrow$ & \textbf{18.36}$\uparrow$ & \textbf{22.11}$\uparrow$ & \textbf{0.94}$\uparrow$ & \textbf{0.77}$\uparrow$ & 0.46 \\
\midrule
HybridMoE 
& Blip2-7B & 0.30 & 0.13 & 0.23 & 0.28 & 23.76 & 12.93 & 13.29 & 15.64 & 0.89 & 0.55 & 0.33 \\
& Paligemma2-10B & 0.45 & 0.20 & 0.34 & 0.35 & 35.97 & 17.25 & 14.20 & 15.94 & 0.89 & 0.59 & 0.32 \\
& Llava-1.5-7B & 0.40 & 0.17 & 0.32 & 0.34 & 32.52 & 16.64 & 12.95 & 14.25 & 0.91 & 0.61 & 0.33 \\
& SmolVLM-Instruct & 0.48 & 0.21 & 0.36 & 0.36 & 50.46 & 24.16 & 16.59 & 20.92 & 0.92 & 0.66 & 0.35 \\
& Gemma3-12b-pt & 0.46 & 0.19 & 0.35 & 0.35 & 38.81 & 17.82 & 11.93 & 14.58 & 0.92 & 0.79 & \textbf{0.55}$\uparrow$ \\
& Qwen2.5-VL-7B-Instruct & \textbf{0.54}$\uparrow$ & \textbf{0.32}$\uparrow$ & \textbf{0.41}$\uparrow$ & \textbf{0.43}$\uparrow$ & \textbf{56.32}$\uparrow$ & \textbf{28.92}$\uparrow$ & \textbf{21.48}$\uparrow$ & \textbf{24.19}$\uparrow$ & \textbf{0.95}$\uparrow$ & \textbf{0.82}$\uparrow$ & 0.47 \\
\bottomrule
\end{tabular}
}
\end{table*}

In contrast, the full HybridMoE framework with IPS injection yields consistent and more comprehensive performance improvements across all evaluation dimensions. The inclusion of IPS acts as a task-specific semantic signal that enhances contextual grounding and idiomatic reasoning, leading to stronger alignment in both textual and multimodal representations. This is reflected in higher BERTScore values (reaching $\approx 0.94$), improved IVS (Idiomatic Validation Score) ($\approx 0.73$--$0.77$), and notable gains in \textit{IDIO-TONE}, indicating better modeling of pragmatic and affective aspects of idiomatic expressions. Additionally, performance across ROUGE and BLEU metrics remains stable or improves marginally, demonstrating that semantic enrichment does not come at the cost of lexical fidelity. Overall, while HybridMoE without IPS primarily strengthens structural fusion and surface-level reasoning, the integration of IPS enables a more balanced and semantically aware system, effectively bridging the gap between lexical accuracy and deeper idiomatic understanding.

This demonstrates that although MoE structures can introduce expert diversity, without IPS, the routing mechanism lacks semantic specificity. Experts are activated based on undifferentiated patterns, leading to diluted idiomatic representation and weaker cross-modal alignment.
\begin{table*}[t]
\centering
\caption{Qualitative Analysis of IDIO-TONE Performance in HybridMoE-Configured Vision-Language Models}
\label{tab:idio_tone_quality}
\scriptsize
\setlength{\tabcolsep}{8pt}
\begin{tabular}{|p{0.46\linewidth}|p{0.46\linewidth}|}
\hline
\begin{center}
\includegraphics[width=0.9\columnwidth]{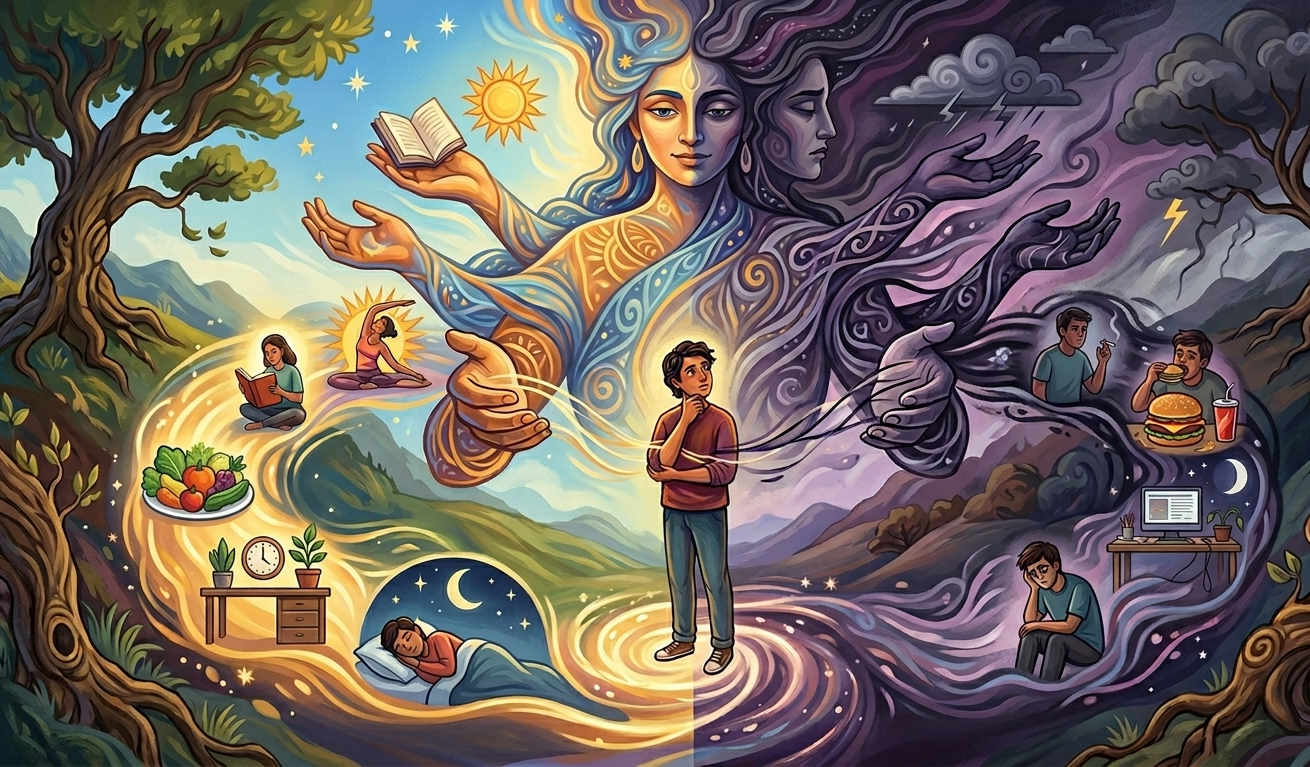}\\
\end{center}
\begin{raggedright}
\textbf{{Idiom:}} \begin{bengali}
    মানুষের অভ্যাসই দেবতা।
\end{bengali}\\
\textbf{{Pronunciation:}} manusher obhyash-i debota\\
\textbf{{Literal Translation:}} A person’s habits are like their God.\\
\textbf{{Idiomatic Understanding:}} This bengali idiom ``\begin{bengali}
    মানুষের অভ্যাসই দেবতা
\end{bengali}'' emphasizes that habit shapes human behavior and character, often becoming so powerful that it governs one’s actions like a higher authority. It reflects the idea that repeated practices strongly influence decision-making and identity, for better or worse.\\
\textbf{Ground Truth:} [`Aspiration']\\
\end{raggedright}
&
\begin{center}
\includegraphics[width=0.9\columnwidth]{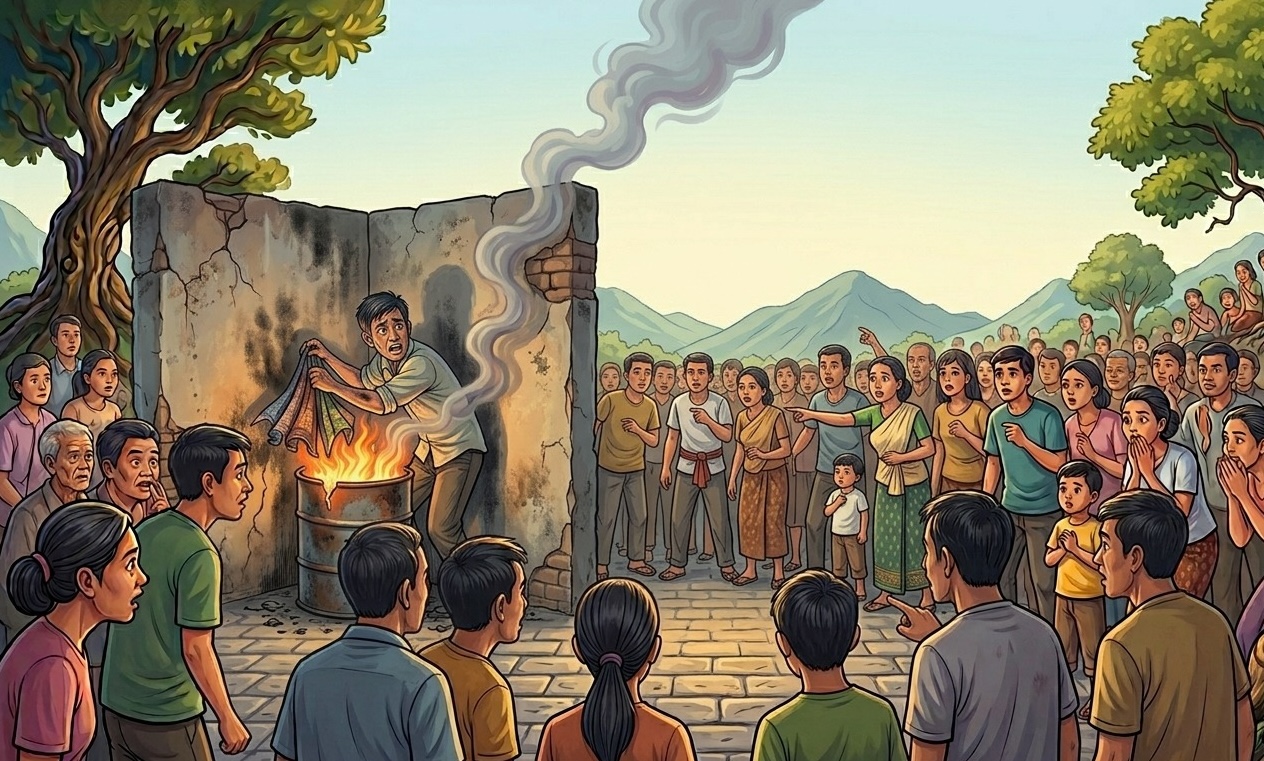}\\
\end{center}
\begin{raggedright}
    \textbf{{Idiom}:} \begin{thai}
    ปิดควันไฟไม่มิด
\end{thai}\\
\textbf{Pronunciation}: pit khwan fai mai mit
\\
{\textbf{Literal Translation:}} You cannot completely hide smoke from a fire.\\
{\textbf{Idiomatic Understanding: }} This Thai idiom ``\begin{thai}
    ปิดควันไฟไม่มิด
\end{thai}''  conveys that the truth cannot be fully concealed, no matter how hard one tries. Just as smoke inevitably reveals the presence of fire, hidden actions, lies, or wrongdoing will eventually become visible. It is often used in contexts involving secrets, deception, or wrongdoing, emphasizing the inevitability of exposure.\\
\textbf{Ground Truth:} [`Deception', `Fear']\\
\end{raggedright}
\\
\hline
\vspace{0.1cm}
\begin{raggedright}
\textbf{Qwen2.5-VL-7B-Instruct: [`Aspiration']} \\[2pt]
SmolVLM-Instruct: [`Affection', `Aspiration']\\
Gemma3-12b-pt : [`Aspiration', `Fear']\\
Paligemma2-10B: [`Aspiration'] \\
Llava-1.5-7B: [`Aspiration']\\
Blip2-7B: [`Aspiration'] \\
\end{raggedright}
&
\vspace{0.1cm}
\begin{raggedright}
\textbf{Qwen2.5-VL-7B-Instruct: [`Deception', `Fear']} \\[2pt]
SmolVLM-Instruct: [`Affection', `Aspiration']\\
Gemma3-12b-pt: [`Fear']\\
Paligemma2-10B: [`Fear', `Sorrow'] \\
Llava-1.5-7B: [`Fear']\\
Blip2-7B: [`Fear'] \\
\end{raggedright}\\
\hline
\end{tabular}
\end{table*}
By contrast, the HybridMoE with IPS explicitly encodes idiomatic abstraction signals into the CLS representations during encoding, enabling task-aware expert routing that prioritizes idiom-grounded context, figurative meaning, and symbolic grounding. This semantic conditioning significantly improves linguistic coherence. Furthermore, idiom-sensitive evaluation via IVS reveals that IPS-enhanced models better capture cultural nuance and metaphorical intent, which are often lost in finetuning or naïve MoE setups. In essence, this ablation confirms that without IPS, models struggle to disambiguate literal vs. figurative content, leading to representational collapse; whereas IPS-enabled HybridMoE transforms multimodal fusion into an idiom-aware, semantically enriched, and performance-consistent mechanism across diverse VLM architectures.

\paragraph{Qualitative Analysis on \textit{IDIO-TONE} Metric.}
As illustrated in Table~\ref{tab:idio_tone_quality}, we conduct a qualitative analysis using the \textit{IDIO-TONE} metric to examine model behavior beyond aggregate scores. The results indicate that VLMs generally align well with the dominant idiomatic tone, while exhibiting variability in capturing secondary and nuanced tonal signals. For the thai idiom \begin{thai} ปิดควันไฟไม่มิด \end{thai} (truth cannot be concealed), the ground truth reflects a combination of \textit{Aspiration} and \textit{Deception}. Most models successfully identify \textit{Fear}, demonstrating strong sensitivity to explicit cues, although they often underrepresent the complementary \textit{Deception} aspect, suggesting challenges in modeling implicit moral or reflective undertones. 

Similarly, for the bengali idiom \begin{bengali}মানুষের অভ্যাসই দেবতা
\end{bengali}, where the ground truth emphasizes \textit{Aspiration}, models consistently capture the primary tone, while occasionally introducing additional labels such as \textit{Fear} or \textit{Affection}. This reflects a tendency toward broader associative reasoning in abstract or philosophical contexts.

Moreover, Qwen2.5-VL-7B-Instruct consistently predicts the correct \textit{IDIO-TONE} labels for both the Bengali and Thai idioms, further demonstrating its superior capability in capturing nuanced idiomatic semantics and pragmatic tone compared to other models. Overall, these observations highlight that while models are effective at identifying dominant tonal categories, capturing multi-label coherence and subtle pragmatic nuances remains an evolving capability. Conclusively, \textit{IDIO-TONE} serves as a crucial and sensitive evaluation metric, enabling a more fine-grained assessment of idiomatic understanding beyond surface-level semantic alignment.

\subsection{LLM usage}
We used large language models (LLMs) to assist with code development and minor editing of the final manuscript.
\end{document}